\documentclass[a4paper,fleqn]{cas-dc}
\pdfoutput=1

\usepackage{hyphenat}
\usepackage{pgf,tikz,pgfplots}
\pgfplotsset{compat=1.14}%1.15
\usepackage{mathrsfs}
\usetikzlibrary{arrows}

\usepackage[numbers]{natbib}

\def\tsc#1{\csdef{#1}{\textsc{\lowercase{#1}}\xspace}}
\tsc{WGM}
\tsc{QE}
\tsc{EP}
\tsc{PMS}
\tsc{BEC}
\tsc{DE}
\newcommand\restr[2]{{% we make the whole thing an ordinary symbol
  \left.\kern-\nulldelimiterspace % automatically resize the bar with \right
  #1 % the function
  \right|_{#2} % this is the delimiter
  }}
\newtheorem{theorem}{Theorem}

\newtheorem{definition}{Definition}
\newtheorem{corollary}{Corollary}
\newtheorem{proposition}{Proposition}
\newdefinition{rmk}{Remark}
\newproof{pf}{Proof}

\usepackage{tikz-cd}
\usepackage[algo2e,ruled,vlined]{algorithm2e}
\DeclareMathOperator{\Sd}{Sd}
\DeclareMathOperator{\st}{st}

\begin{document}
\let\WriteBookmarks\relax
\def\floatpagepagefraction{1}
\def\textpagefraction{.001}
%\usepackage{tikz}
%\usetikzlibrary{matrix}
%\usepackage{tikz-cd}
\shorttitle{Universal approximation theorem by simplicial approximation}
\shortauthors{Paluzo-Hidalgo E., Gonzalez-Diaz R., Guti\'errez-Naranjo M. A.}

\title [mode = title]{Two-hidden-layer Feed-forward Networks are Universal Approximators: a Constructive Approach 
%by means of Simplicial Complexes
}                      
%\tnotemark[1,2]

%\tnotetext[1]{This document is the results of the research
%   project funded by the National Science Foundation.}

%\tnotetext[2]{The second title footnote which is a longer text matter
%   to fill through the whole text width and overflow into
%   another line in the footnotes area of the first page.}

\author[1]{Eduardo Paluzo-Hidalgo}[orcid=0000-0002-4280-5945]
\cormark[1]
%\fnmark[1]
\ead{epaluzo@us.es}
\ead[URL]{https://personal.us.es/epaluzo}
\author[1]{Rocio Gonzalez-Diaz}[orcid=0000-0001-9937-0033]
%\cormark[1]
%\fnmark[1]
\ead{rogodi@us.es}
\ead[url]{https://personal.us.es/rogodi}
\address[1]{Department of Applied Mathematics I, School of engineering, University of Seville, Seville, Spain}

\author[2]{Miguel A. Guti\'errez-Naranjo}[orcid=0000-0002-3624-6139]
\ead{magutier@us.es}
\ead[URL]{http://www.cs.us.es/~naranjo/}
\address[2]{Department of Computer Science and Artificial Intelligence, School of engineering, University of Seville, Seville, Spain}

\cortext[cor1]{Corresponding author}
%\cortext[cor2]{Principal corresponding author}
%\fntext[fn1]{This is the first author footnote. but is common to third
%  author as well.}
%\fntext[fn2]{Another author footnote, this is a very long footnote and
%  it should be a really long footnote. But this footnote is not yet
%  sufficiently long enough to make two lines of footnote text.}

\begin{abstract}
It is well-known that artificial neural networks are universal approximators. The classical existence result proves that, given a continuous function on a compact set 
embedded in
an $n$-dimensional space, 
there exists a one-hidden-layer feed-forward network that
approximates the function. 
In this paper, a constructive approach to 
this problem
is given for the case of a
continuous function on triangulated spaces.
Once a triangulation of the space is given, a 
two-hidden-layer feed-forward network with a concrete set of weights
is computed.
The 
level
of the approximation depends on the refinement of the 
triangulation.
\end{abstract}

%\begin{graphicalabstract}
%\includegraphics[width = 0.8\linewidth]{pagina.pdf}
%\end{graphicalabstract}

%\begin{highlights}
%\item Constructive approach to the Universal Approximation Theorem.
%\item Relationship between neural networks and simplicial approximations is given.
%\item The level of the approximation depends on the refinement of the triangulation.
%\end{highlights}


\maketitle

%%%%%%%%%%%%%%%%%%%%%%%%%%%%%%%%%%%%%%%%%%%%%%%%%%%%%%%%%%%%
%%%%%%%%%%%%%%%%%%%%%%%%%%%%%%%%%%%%%%%%%%%%%%%%%%%%%%%%%%%%

%%
\section{Introduction}
One of the first results in the development of neural networks is 
the {\it Universal Approximation Theorem}
\cite{DBLP:journals/mcss/Cybenko89, Hornik:1991:ACM:109691.109700}. This classical result shows that any  continuous function on a compact set in ${\mathbb R}^n$ can be approximated by a multi-layer feed-forward network with only one hidden layer and non-poly\-nomial activation function (like the sigmoid function).
It is well-known that 
this result has
two important drawbacks for 
its practical use: firstly,
the width of the hidden layer grows exponentially and, 
secondly,
the proofs developed in \cite{DBLP:journals/mcss/Cybenko89, Hornik:1991:ACM:109691.109700}
do not provide a practical algorithm for building such a network.
Bearing these results in mind, many researchers are paying attention to theoretical aspects of the current success of neural network architectures and searching
for bounds for the depth and width of such networks and the possibility 
that they act
as universal approximators (see, e.g., \cite{DBLP:journals/corr/LiangS16,pmlr-v70-safran17a,DBLP:journals/corr/Telgarsky16} among many others).

Undoubtedly,
the use of many hidden layers is a big contribution to the success of deep learning architectures \cite{DBLP:conf/aaai/SunCWLL16}, but instead of exploring the power of depth, recently several studies have made interesting contributions 
about
the power of width \cite{1710.11278,DBLP:conf/nips/LuPWH017,DBLP:journals/corr/abs-1803-00094}. To sum up, 
these works
show that there exist continuous functions on compact sets that
cannot be approximated by  
any
neural network if the width of the layers is not larger than a bound, regardless of the depth of the network. In \cite{GULIYEV2018296}, the authors constructively proved  that single-hidden-layer feed-forward networks with fixed weights are universal approximators for univariate functions, and they provided a step-by-step construction. However, as the authors claimed, 
not all continuous multivariate functions can be approximated by such neural networks. Their case of study could be considered somehow a particular case of our approach for the 1-dimensional case. 

Other interesting research line on the expressive power of neural networks follows an algebraic approach. In \cite{DBLP:conf/nips/DelalleauB11,MartensM14, poon2012sumproduct}, sum product networks are explored and tensor properties are studied in \cite{DBLP:conf/colt/CohenSS16,DBLP:conf/icml/CohenS16}. A different perspective was introduced in \cite{DBLP:conf/nips/KileelTB19} where "deep polynomial neural networks" (where the activation function is a polynomial exponentiation) are considered.

In this paper, we provide an effective method for finding the weights of a two-hidden-layer
feed-forward network which approximates 
a given continuous 
function between two triangulable 
metric spaces.
Let us remark that the method is constructive, and it only depends on the desired level of approximation to the 
given function.
Our approach is based on a classical result from algebraic topology. Roughly speaking, our result is 
based on
two observations:
Firstly, triangulable 
spaces
can be ``modelled'' 
using
simplicial complexes, and a continuous 
function between two triangulable spaces can be approximated by a simplicial map
between simplicial complexes. Secondly,
a simplicial map between simplicial complexes can be ``modelled'' as a two-hidden-layer feed-forward network.
Let us remark that the classical result Universal Approximaton Theorem is valid for all compact set on ${\mathbb R}^n$ and our 
results presented here are valid for triangulable 
 spaces.
However, triangulable 
 spaces
are  
common in real-world problems. 
Furthermore, we would like to highlight that an advantage of our approach from a theoretical and practical point of view is that it can be useful to solve real-world problems that can be modeled by triangulable spaces.

The paper is organized as follows: In Section \ref{sec:background}, the preliminary notions about multi-layer feed-forward networks and simplicial complexes are provided. Then, in Section \ref{sec:MLPandSM}, a concrete architecture 
of such networks
that acts equivalently to a given simplicial map is given. In Sections \ref{sec:SimpAprox} and \ref{sec:UnivApprox}, we extend the 
Simplicial Approximation Theorem and Universal Approximation Theorem, respectively. The complexity of the architecture is studied in Section 6.
A specific example is described in Section \ref{sec:example}. Finally, conclusions are given in Section \ref{sec:conclusions}.

%%%%%%%%%%%%%%%%%%%%%%%
%% BACKGROUND %%%%
%%%%%%%%%%%%%%%%%%%%%%%
\section{Background}\label{sec:background}

In this section, we recall some notions about artificial neural networks and simplicial complexes.

\subsection{Multi-layer 
Feed-forward Networks }

Artificial neural networks are 
inspired 
by 
biological networks of alive neurons in a brain. The number of different architectures, algorithms, and areas of application has recently grown in many directions. In general, a neural network can be formalized as a 
function
${\cal N}_{\omega,\Theta}: {\mathbb R}^n \to {\mathbb R}^m$ 
that
depends on a set of weights
$\omega$ and a set of parameters $\Theta$ 
which involves the description of activation functions, layers, synapses between nodes (neurons), and whatever other consideration in its architecture. A good introduction to artificial neural networks can be found in \cite{haykin99a}.

In this paper, we focus on one of the simplest classes of artificial neural networks:  multi-layer feed\-for\-ward networks. They
consist of three or more  fully connected layers of nodes:
an input layer, an output layer, and one or more hidden layers. Each node in one layer has an 
activation function and it 
is connected with
every node in the following layer. The next definition formalizes this idea.

\begin{definition}[adapted from \cite{Hornik:1991:ACM:109691.109700}]\label{def:ANN}
A multi-layer feed-forward network defined on a real-valued $n$-dimensional space is a 
function
${\cal N}: {\mathbb R}^n \to {\mathbb R}^m$  such that, for each $x \in {\mathbb R}^n$, ${\cal N}(x)$ is 
the composition of
$k+1$
functions
$${\cal N}(x) =  
f_{k+1}
\circ f_k\circ \dots 
 \circ   f_1  (x)$$
where $k\in\mathbb{Z}$ is the number of hidden layers,  
$k\geq 1$, and, for  $1\leq i\leq k+1$,
 $f_i:  {\mathbb R}^{d_{i-1}} \to {\mathbb R}^{d_i}$  
 is defined as 
  $$f_i(y) =\phi_i(W^{(i)}; y;b_i)$$ being
 $W^{(i)}$
 a real-valued $d_{i-1}\times d_i$ matrix,
 (that is, $W^{(i)}\in {\cal M}_{d_{i-1}\times d_i}$),
 $b_i\in \mathbb{R}^{d_i}$  the bias term, and $\phi_i$ 
 a bounded,  continuous, and non-constant function (called activation function). 
 Notice that $d_0=n$, $d_{k+1}=m$ and $d_i\in {\mathbb Z}$, $1\leq i\leq k$, is called the width of the $i$-th  hidden layer. 
\end{definition}

Next, we rewrite one of the most important theoretical results of multi-layer feed-forward networks adapted to our notation.

\begin{theorem}[Universal Approximation Theorem, \cite{Hornik:1991:ACM:109691.109700}]\label{th:unaprox}
Let $A$ be any compact subset of $\mathbb{R}^n$ 
and let $C(A)$ be the space of real-valued continuous functions on $A$.
Then, given any $\epsilon>0$ and any function $g\in C(A)$, 
there exists a multi-layer feed-forward network ${\cal N}:\mathbb{R}^n\to\mathbb{R}$
approximating $g$,  that is,
$||g-{\cal N}||<\epsilon $. 
\end{theorem}

As far as we know, the existing
proofs 
of
this theorem are non-constructive. See \cite{hornik89a,Hornik:1991:ACM:109691.109700,DBLP:journals/mcss/Cybenko89}
where it is claimed 
that 
there exists a one-hidden-layer feed-forward network ${\cal N}:\mathbb{R}^n\to\mathbb{R}$, defined as ${\cal N}(x)=f_2\circ f_1 (x)$ with $f_1(y)= \phi_1 (W^{(1)};$ $y;b_1)$ and $f_2(y)=W^{(2)} y$, 
approximating 
$g$, that is,
$||g-{\cal N}||<\epsilon $,
but  no general algorithm to build ${\cal N}$ is given. 
In this paper, we will provide a constructive approach to
Theorem \ref{th:unaprox} through a two-hidden-layer feed-forward network.
The most important restriction to our approach is that our result is valid for triangulable spaces
instead of compact ones but, as pointed out above, 
triangulable spaces cover
most of the real-world problems.

\subsection{Simplicial Complexes}
In this subsection, we recall the main result used in this paper based on a 
classical theorem from algebraic topology
known as the {\it Simplicial Approximation Theorem}. 
For a further comprehension of the field, \cite{ayala2002elementos,boissonnat2018geometric,edelsbrunner2010,Hatcher,munkres} can be consulted.

Simplicial complexes are a data structure widely used to represent topological spaces. They are 
versatile mathematical objects
that decompose a given
topological space into pieces 
called simplices.

\begin{definition}
Let $v_0,v_1,\dots,v_i$ be  $i+1$ affinely independent points in $\mathbb{R}^n$ 
(being $i\leq n$).
An $i$-simplex, $\sigma = (v_0,v_1, \dots, v_i)$, is the convex set
$$\Big\{x\in \mathbb{R}^n \ \big | \ x= \sum_{j=0}^i \lambda_j v_j \ \mbox{ with } \ \lambda_j\ge 0
\mbox{ and } \sum_{j=0}^i \lambda_j=1
\Big\}. $$
The points $v_0,v_1,\dots,v_i$ are the vertices of $\sigma$. 
The dimension of $\sigma$
is $i$.
We say that 
$\sigma'$ is a face of $\sigma$
(denoted as 
$\sigma'\preceq \sigma$)
if $\sigma'$ is an $i'$-simplex, with $i'\leq i$, and whose vertices are also vertices of $\sigma$. 
\end{definition}

When several simplices are joint together, a more complex structure called simplicial complex is built. The following definition exhibits the way simplices can be glued together to obtain such a simplicial complex. 

\begin{definition}
    A     simplicial complex $K$ is a 
    collection of simplices  such that:
    \begin{enumerate}
        \item $\sigma\in K$ and 
        $\sigma'\preceq \sigma$
        implies $\sigma'\in K$;
        \item
        $\sigma$, $\mu\in K$ implies $\sigma\cap \mu$ is either empty or a face of both.
    \end{enumerate}
If such collection is finite, then $K$ is a finite simplicial complex.    
\end{definition}

  The  underlying space of $K$, denoted by $|K|$, is the union of the simplices of $K$ together with the topology inherited from the ambient Euclidean space where the simplices are placed.
  
  \begin{definition}
    A simplex $\sigma \in K$ is maximal if it is not a face of any other simplex in $K$.
The dimension of $K$, denoted by $\dim(K)$  is the largest  of the dimensions of its maximal simplices. The simplicial complex $K$ is pure if  
    all the maximal simplices have  dimension equal to $\dim(K)$.
    \end{definition}

  A subcomplex $L$ of $K$ is a simplicial complex such that $L\subseteq K$.
  The skeleton of $K$ is a particular subcomplex of $K$. Let us remark that the 0-skeleton of $K$ is its vertex set.
  
  \begin{definition}
  The subcomplex of $K$ consisting of all the simplices of $K$ of dimension $j$ or less is called the $j$-skeleton of $K$ and it is denoted by $K^{(j)}$. 
    \end{definition}

Next, we recall 
the concept of the star of a simplex $\sigma$ in a simplicial complex $K$. Intuitively, it is the subcomplex of $K$ whose maximal simplices share $\sigma$ as a face.

\begin{definition}
    Let $K$ be a simplicial complex and $\sigma$ a simplex
    of $K$. The star of $\sigma$ in $K$, denoted by $\st (\sigma)$, is defined as:
   $$\st (\sigma) = \big\{\mu \in K \ \big| \ \exists \ \xi \in K \mbox{ such that } \sigma \preceq \xi \mbox{ and } \mu\preceq \xi \big\}.
   $$
\end{definition}

As said above, simplicial complexes are combinatorial data structures used to model topological spaces. A way to obtain a refined model from an existing one is to subdivide  it into small pieces so
that the result is  topologically equivalent to the 
former.
 
\begin{definition}
Let $K$ and $K'$ be simplicial complexes. It is said that $K'$ is a subdivision of $K$ if:
\begin{enumerate}
    \item $|K|=|K'|$;
    \item     $\sigma' \in K'$ implies that there exists $\sigma\in K$ such that $\sigma'$
    is contained in
    $\sigma$.
\end{enumerate}
\end{definition}

The barycentric subdivision is a concrete example of subdivision of simplicial complexes.
Using that the barycenter of an
 $i$-simplex $\sigma= (v_0\dots v_i)$ is $$b(\sigma) = \sum_{j=0}^i\frac{1}{j+1}v_j,$$
the definition
of barycentric subdivision of a simplicial complex arises in a natural way.

\begin{definition}
Let $K$ be a simplicial complex. The
barycentric subdivision of the $0$-skeleton of $K$ is defined as the set of vertices of $K$, that is, $\Sd K^{(0)} = K^{(0)}$. Assuming
we have $\Sd K^{(i-1)}$, which denotes the barycentric subdivision of the $(i-1)$-skeleton of $K$, $\Sd K^{(i)}$ is built by adding the barycenter of every $i$-simplex as a new vertex and connecting it to the simplices that subdivide the boundary of such $i$-simplex.
The barycentric subdivision of $K$, denoted by $\Sd K$, is $\Sd K^{(u)}$ where $u$ is the dimension of $K$.
The iterated application of barycentric subdivisions is denoted by $\Sd^t K$ where $t$ is the number of iterations.
\end{definition}

\begin{figure}
    \centering
    \includegraphics[width =0.8 \linewidth]{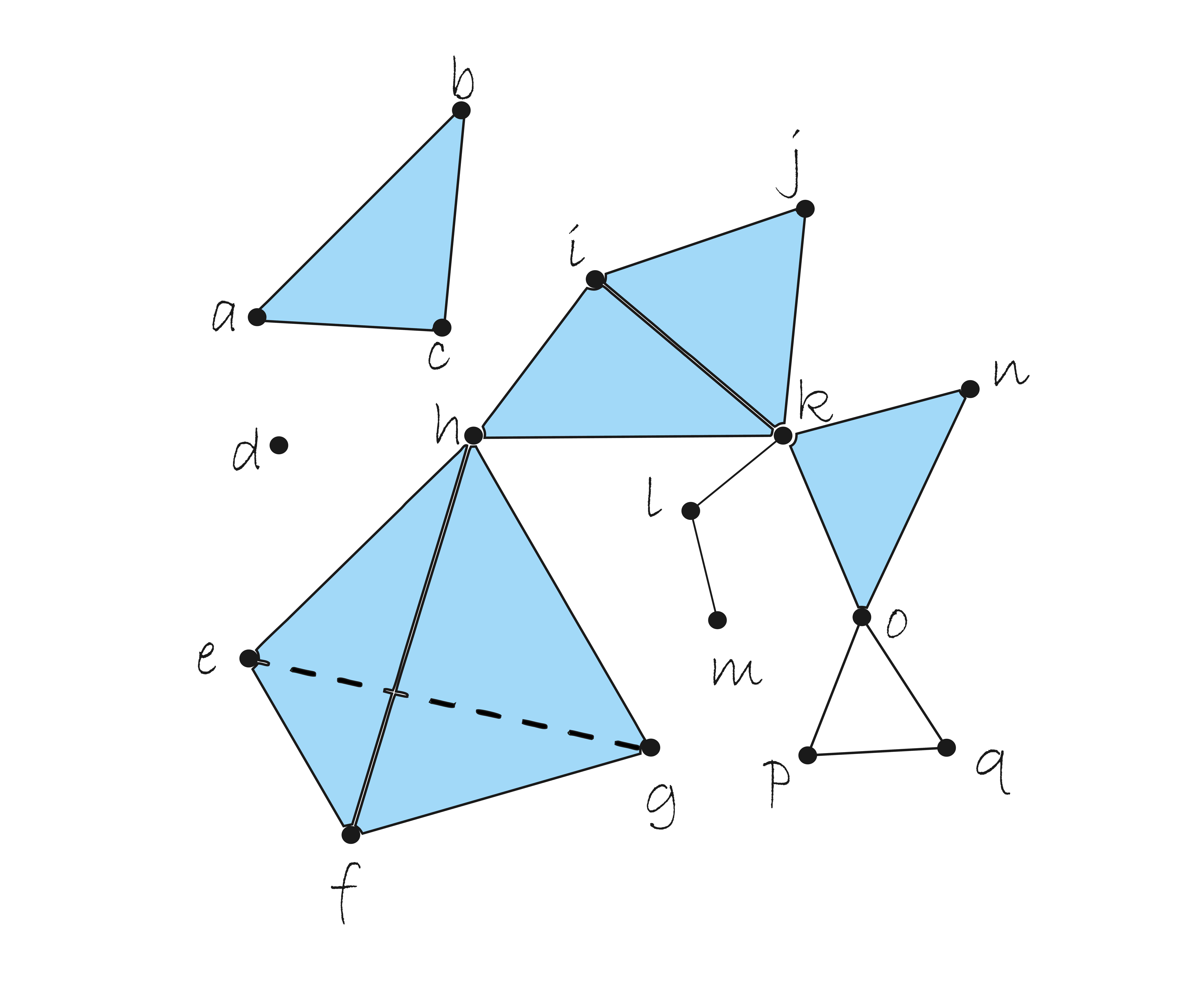}
    \caption{Example of a simplicial complex. $(d)$ is a $0$-simplex, $(o,p)$ is a $1$-simplex , $(a,b,c)$ is a $2$-simplex, and $(e,f,g,h)$ is a $3$-simplex. The $1$-simplex $(a,b)$ is a face of    
    $(a,b,c)$. The maximal simplices of the simplicial complex are
    $(e,f,g,h)$, $(a,b,c)$, $(i,h,k)$, $(i,j,k)$, $(o,p)$, $(o,q)$, $(p,q)$, and $(d)$. The dimension of    the simplicial complex
    is 3.}
    \label{fig:simplicial_complex}
\end{figure}

\begin{figure}
    \centering
    \includegraphics[width =0.8 \linewidth]{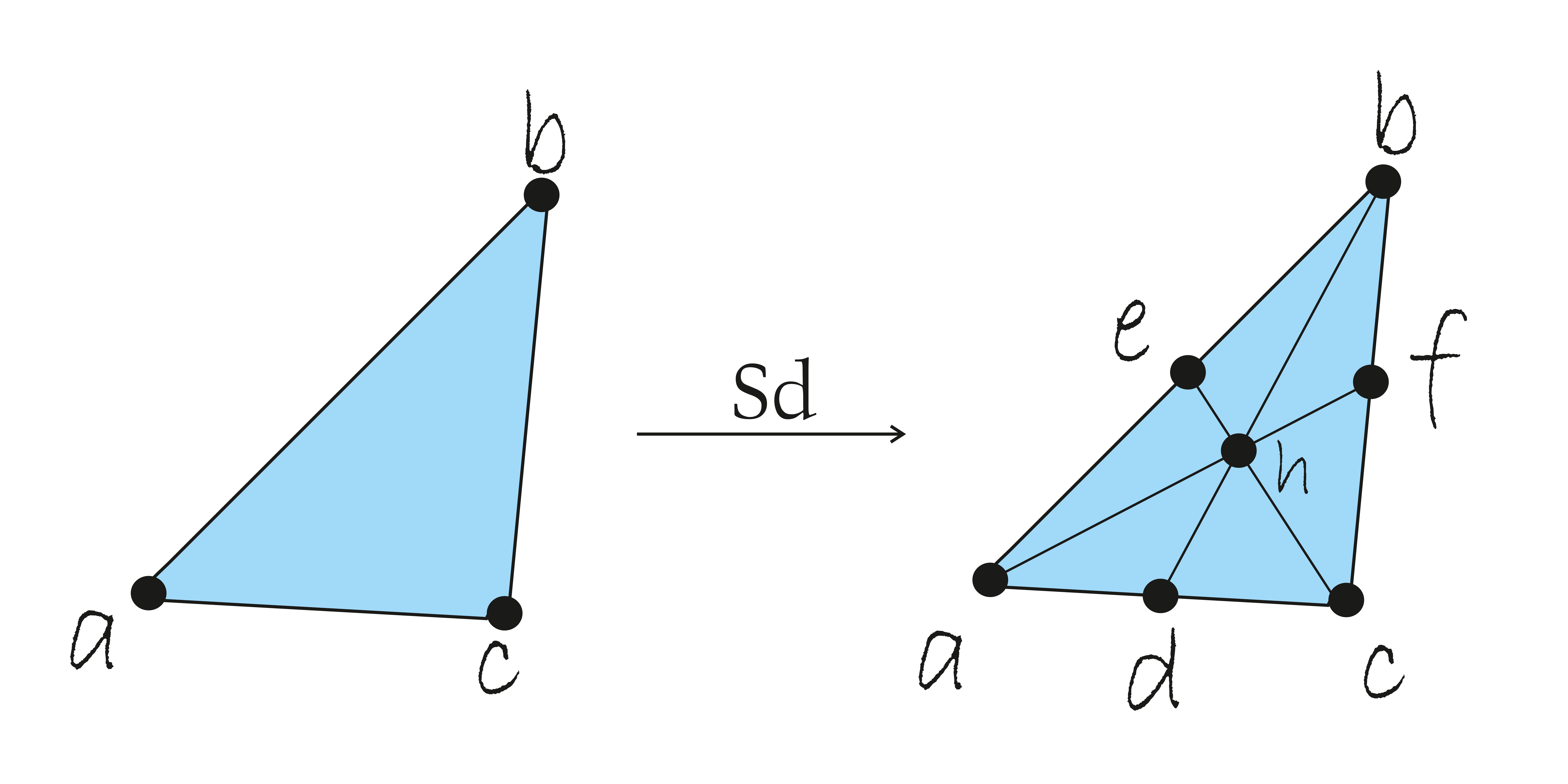}
    \caption{On the left, a simplicial complex with just one maximal simplex, the $2$-simplex $(a,b,c)$. 
    On the right, its first barycentric subdivision.
    }
    \label{fig:barycentric_sub}
\end{figure}

Let us see now how the ``geometric size'' of the simplices of a simplicial complex can be measured.

\begin{definition}
    Let $K$ be a  finite simplicial complex. The diameter of a simplex $\sigma$ in $K$ is defined as 
    $$\delta(\sigma) =    \max 
    \big\{ ||x-y|| \ \mbox{ such that } \ x,y
    \mbox{ are vertices of }
    \sigma\big\}$$ 
        and  the  mesh of $K$ is defined as 
    $$m(K) = \max 
    \big\{\delta(\sigma) \ \mbox{ such that } \ \sigma \in K \big\}. $$
\end{definition}

\begin{theorem}[{\cite[p. 86]{munkres}}]\label{th:barycentricepsilon}
 Given a simplicial complex $K$ and  a real number $\epsilon>0$, there exists an integer  $t>0$ such that $m(\Sd^t K)\le \epsilon$.
\end{theorem}

Let us now think about maps between simplicial complexes. These maps can be considered as extensions of simpler maps defined 
between the corresponding vertices of two given simplicial complexes. Interestingly, such maps can be considered as approximations of continuous functions defined on the underlying topological spaces that the simplicial complexes are modeling. Let us formalize these notions.

\begin{definition}
    Let $K$ and $L$ be two simplicial complexes. A vertex map is a 
    map
     $\varphi:K^{(0)}\to L^{(0)}$
     with the property that 
     for every simplex $\sigma$ in $K$ there exists a simplex $\mu$ in $L$ such that the vertices of $\sigma$ map to vertices of $\mu$.
\end{definition}

A vertex map $\varphi$ can be extended to a continuous function $\varphi_c:|K|\to |L|$ in the following way.

\begin{definition}
Let $K$ and $L$ be two  simplicial complexes and let  $\varphi:K^{(0)}\to L^{(0)}$ be a vertex map.
The simplicial map $\varphi_c$  induced by $\varphi$ is defined as follows.
Let $x\in |K|$. Then there exists an $i$-simplex  $\sigma=(v_0,\dots,v_i)$ in $K$ and numbers 
    $\lambda_j\ge 0$ 
    such that
     $$\sum_{j=0}^i \lambda_j=1\,\mbox{ and }\,
 x= \sum_{j=0}^i 
 \lambda_j v_j.$$ Then
$$ \varphi_c(x) = \sum_{j=0}^i  \lambda_j  \varphi(v_j).$$
\end{definition}

Any vertex map $\varphi$ induces a simplicial map $\varphi_c$, but if we want that 
map to be a simplicial
approximation of a continuous function
between the underlying spaces of
two simplicial complexes $K$ and $L$, a restriction
on the star of each vertex of $K$ must
be added.

\begin{definition} 
    Let $K$ and $L$ be simplicial complexes and  $g:|K|\to |L|$ a continuous 
    function.
    A simplicial map $\varphi_c: |K|\to |L|$ induced by a vertex map $\varphi: K^{(0)} \to L^{(0)}$ is a simplicial approximation of $g$ if $$g(|\st (v)\ |)\subset |\st (\varphi(v))\ | $$
    for each vertex $v$ of $K$.
\end{definition}

The Simplicial Approximation Theorem 
ensures the existence of simplicial maps that approximate continuous functions as close as we want.

\begin{theorem}
[Simplicial Approximation Theorem {\cite[p.  56]{edelsbrunner2010}}]
\label{th:simaprox}
If $g:|K|\to |L|$ is continuous then there is a sufficiently large integer $t>0$ such that $\varphi_c:|\Sd^t K|\to |L|$ is a simplicial approximation of $g$.
\end{theorem}

Theorems \ref{th:unaprox} and \ref{th:simaprox} 
are key results in their respective fields. Our aim in this paper is to consider Theorem \ref{th:simaprox} as a pillar to obtain a 
constructive approach to
Theorem \ref{th:unaprox}. Roughly speaking, the idea is to consider a simplicial approximation 
of a continuous function
between two simplicial complexes.
Such a simplicial approach is characterized through
a vertex map which can be expressed as a neural network. 
Besides,
the simplicial approximation can be chosen in such a way that it approximates the continuous function
between the underlying spaces of two simplicial complexes. Since the parameters of the neural network can be effectively obtained from the vertex map, this method provides a constructive way to find a neural network
that
approximates a continuous function between
the underlying spaces of two simplicial complexes.
Following that aim, let us formalize the relationship between topological spaces and simplicial complexes.

\begin{definition}
    A triangulation of a topological space $X$ consists of a simplicial complex $K$ and a homeomorphism $\tau:X\to |K|$.
    We say that the triangulation is finite if $K$ is finite. We say that $X$ is (finitely) triangulable if such (finite) triangulation exists.
\end{definition}

\begin{figure}
    \centering
    \includegraphics[width = 0.4 \linewidth]{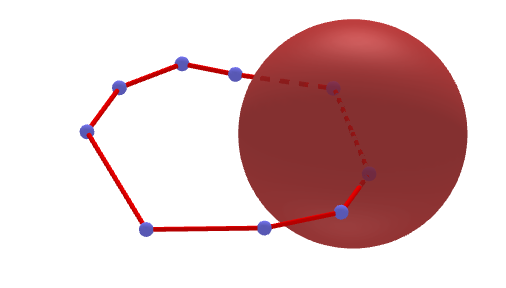}
    \includegraphics[width = 0.4 \linewidth]{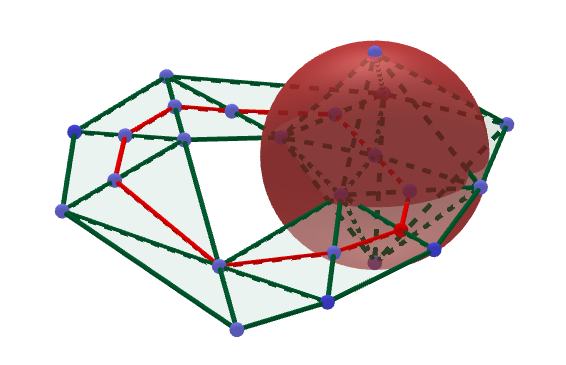}
    \includegraphics[width = 0.4 \linewidth]{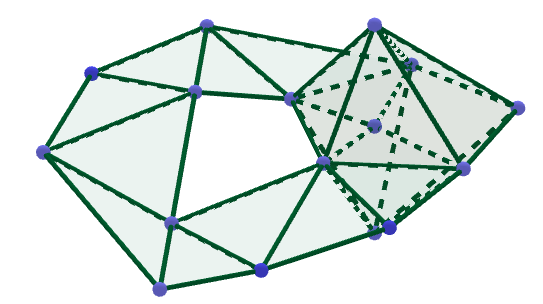}
    \caption{From left to right and from top to bottom: (1) A sphere and a loop that intersect; (2) the sphere and the loop with a possible triangulation (the loop is covered by $2$-simplices); (3) the triangulation of the loop and the sphere.}
    \label{fig:triangulation_maximal_simplices}
\end{figure}

The spaces that can be triangulated by simplicial complexes (see \cite[Theorem A.7, p. 525]{Hatcher}) are 
compact, locally contractible spaces
that can be embedded in $\mathbb{R}^n$ for some $n$. Let us remark that the Universal Approximation Theorem (Theorem \ref{th:unaprox}) is valid for any compact subset of ${\mathbb R}^n$, regardless 
of whether
they are locally contractible or not. Not all compact sets in a metric space are locally contractible (see \cite[Chapter 17.7, p. 426]{geoghegan2010topological}).
Nevertheless, as far as we know, non-locally contractible spaces are very odd in ${\mathbb R}^n$ and this technical topological property has no practical application in real-world problems. Therefore, we can say that the results proved on triangulable spaces are true on a large amount of current neural network problems. Besides, 
we will restrict ourselves to finite pure 
triangulations. 
The other cases could be obtained using homotopies. 
For example, given a sphere and a circumference that intersect transversely, the sphere can be triangulated  using $2$-simplices, and the circumference can be covered by $2$-simplices obtaining a finite pure triangulation of a space homotopic to the initial one (see Figure \ref{fig:triangulation_maximal_simplices}). 

Next, we  extend the definition of a mesh of a simplicial complex in the following way.

\begin{definition}
    Let $X$ be a  triangulable  
    metric
    space and $(K,\tau)$ a finite triangulation of $X$. The mesh of $X$ induced by $(K,\tau)$ is defined as 
    $$\tilde{m}_{(K,\tau)}(X) = \max \big\{\tilde{\delta}(\sigma) \ | \ \sigma \in K \big\} $$
    where 
  $$\tilde{\delta}(\sigma) = \max \big\{   d_X(x,y)
  %||x-y||
  \ | \ x = \tau^{-1}(a), \ y = \tau^{-1}(b);  
  a,b\in \sigma
  \big\}
  $$
    is the extended diameter of a simplex.
\end{definition}

%%%%%%%%%%%%%%%%%%%%%%%%%%%%%%%
%% MAIN RESULTS %%%%%%%%%%%%%%%
%%%%%%%%%%%%%%%%%%%%%%%%%%%%%%%

\section{Multi-layer Feed-forward Networks and Simplicial Maps}\label{sec:MLPandSM}

In this section, we will show that 
simplicial maps can be modelled via multi-layer feed-forward networks in a straightforward way. 

In the following theorem, we will compute a two-hidden-layer feed-forward network to
model a simplicial map $\varphi_c: |K|\to |L|$ where 
$K$ and $L$ are
finite pure simplicial complexes.  
This is not an important constraint in our case, since our final aim is to design a multi-layer feed-forward network that approximates a continuous function between finitely triangulable spaces.

\begin{theorem}\label{th:simplicialmap_nn}
Let us consider
 a simplicial map $\varphi_c:|K|\to |L|$ 
 between the underlying space of two finite pure simplicial complexes $K$ and $L$.
 Then a two-hidden-layer feed-forward network ${\cal N}_{\varphi}$ 
  such that $\varphi_c(x) = {\cal N}_\varphi(x) $ for all $x\in |K|$ can be explicitly defined.
\end{theorem}

\begin{pf} 
Let us assume that $\dim(K)=n$ and $\dim(L)=m$.
Besides, 
let $\{\sigma_1,\dots \sigma_k\}$ be the maximal $n$-simplices of $K$,  where $\sigma_i = \big[v_0^i,\dots,v_n^i\big]$ for all $i$; and let $\{\mu_1,\dots,\mu_{\ell} \}$ be the maximal $m$-simplices of $L$,
 where $\mu_j = \big[u_0^j$, $\dots$, $u_m^j\big]$ for all $j$. 
Let us consider a multi-layer feed-forward network ${\cal N}_\varphi$ with the following architecture: 
\\
(1) An input layer composed by $d_0=n$ neurons; 
\\
(2) a first hidden layer composed by $d_1=k\cdot (n+1)$ neurons; 
\\
(3) a second hidden layer composed by $d_2=\ell \cdot (m+1)$ neurons; and
\\
(4) an output layer with $d_3=m$ neurons. 
\\
Then, let ${\cal N}_\varphi = f_3\circ f_2\circ f_1$ being $f_i (y)= \phi_i (W^{(i)};y;b_i)$, $i=1,2,3$. 
Now, the idea  is to encode
the simplicial complexes involved in the mapping in the hidden layers of the multi-layer feed-forward network. Firstly, a point $x$ in ${\mathbb R}^n$ is transformed into a $k \cdot (n+1)$ vector. This vector can be seen as the juxtaposition of $k$ vectors of dimension $n+1$, one for each of the $k$ simplices in $K$. Each vector of dimension $n+1$ represents the barycentric coordinates of $x$ with respect to the corresponding simplex. 
The matrix $W^{(1)}\in {\cal M}_{k(n+1)\times n }$ and the bias term $b_1$ can be obtained from the barycentric coordinates relations
as follows.
Firstly, 
$$
W^{(1)} = 
\begin{pmatrix}
W^{(1)}_1 \\ \vdots \\ W^{(1)}_k
\end{pmatrix}
$$
where $W^{(1)}_i\in {\cal M}_{(n+1)\times n}$ 
is:
$$
\begin{pmatrix}
v_0^i & \cdots & v_n^i \\
1 & \cdots & 1
\end{pmatrix}^{-1}
=
\begin{pmatrix}
W^{(1)}_i & \big| &
B_i
%A_i & \big| & B_i
\end{pmatrix}
$$
being $\{v_0^i, \dots, v_n\}$ the set of vertices of the maximal simplex $\sigma_i$ of $K$, and second, $b_1 \in \mathbb{R}^{n+1}$ is:
$$
b_1 = \begin{pmatrix}
B_1 \\ \vdots \\ B_k
\end{pmatrix}.
$$
The function $\phi_1$ is then defined as:
 $$\phi_1 (W^{(1)};y;b_1)=W^{(1)}y+b_1.$$

The matrix of weights $W^{(2)}\in {\cal M}_{\ell(m+1)\times k(n+1)}$ between the first and the second hidden layer of ${\cal N}_\varphi$
encodes the vertex map
$\varphi$.
The first hidden layer is composed by $k \cdot (n+1)$ neurons that correspond to the vertices of the maximal simplices of $K$.
The second hidden layer is composed by  $\ell \cdot (m+1)$ neurons that correspond to the vertices of the maximal simplices of $L$.
The matrix $W^{(2)}$ is composed by values zeros and ones. On the one hand, an element of $W^{(2)}$ has value 1 if 
the corresponding vertices 
in
$K$ and $L$ are related by
the vertex map $\varphi$, and, on the other hand, it has  value 0 if
they are not related by $\varphi$. Then,  
$$
W^{(2)} = \big(W^{(2)}_{s_1,s_2}
\big)
$$ 
where$$    W^{(2)}_{s_1,s_2} = \left\{ \begin{array}{l} 1 \mbox{ if } \varphi (v^i_t)=u^j_r ,
\\ 0   \mbox{ otherwise;} \end{array}\right.$$
being $s_1 = j(r+1)$ and $s_2 = i(t+1)$ for $i=1,\dots, k$; $j = 1,\dots, \ell$; $t =0,\dots,n $; and  $r = 0,\dots, m$.
The bias term $b_2$ is the null vector.
Then, the function $\phi_2$ is defined as:
$$\phi_2 (W^{(2)};y;b_2)=W^{(2)}y.$$
The output of the second hidden layer can be seen as the juxtaposition of $\ell$ vectors of dimension $m+1$, one vector for each simplex in the simplicial complex $L$. Each of these vectors represents the barycentric coordinates of $\varphi_c(x)$ with respect to the corresponding simplex in $L$. 
In the next step, only vectors whose all
coordinates are greater than or equal to zero are considered. This condition encodes the simplices of $L$ to which $\varphi_c(x)$ belongs. 
Then, $\phi_3(W^{(3)};y;b_3)$ maps
the barycentric coordinates of $\varphi_c(x)$ with respect to each maximal simplex of $L$ to which $\varphi_c(x)$ belongs, to the Cartesian coordinates of $\varphi_c(x)$. Specifically,  
$$
W^{(3)} = 
\begin{pmatrix}
W^{(3)}_1& \cdots & W^{(3)}_\ell
\end{pmatrix}\in {\cal M}_{m\times \ell(m+1) },
$$
being $W^{(3)}_j=
\begin{pmatrix}
u_0^j & \cdots & u_m^j
\end{pmatrix}$; and
 $b_3$ is the null vector.
Finally,
$\phi_3$ is defined as:
%%%%%%%%%%%%%%%%%%
$$\phi_3 (W^{(3)};y;b_3)=\frac{\sum_{j=1}^{\ell}z^j\psi(y^j)}{\sum_{j=1}^{\ell}\psi{(y^j)}}$$
for
$y = 
\begin{pmatrix}
y^1 \\
\vdots \\
y^{\ell}
\end{pmatrix}\in {\cal M}^{\ell\cdot (m+1)}
$,
with
$z^j = W^{(3)}_j y^j$ and  $\psi(y^j) = 1$ if all the coordinates of $y^j$ are greater than or equal to $0$ and $\psi(y^j)=0$ otherwise.

The 
particular choice
of $\phi_3$ and $\psi$ is motivated by the use of the barycentric coordinates. 
Let us observe that the barycentric coordinates vary with respect to the maximal simplex considered to compute them and that the barycentric coordinates are computed from the  coordinates of the vertices of the maximal simplex considered. Besides, maximal simplices can share common vertices. Then, the map
$\psi$ is used to determine if a given input is located in a specific simplex. The map $\phi_3$ is used to normalize the result in case that a point belongs to more than one simplex.
\hfill \qed
\end{pf}

Summing up, Theorem \ref{th:simplicialmap_nn} establishes that a two-hidden-layer feed-forward network can act equivalently to a simplicial map. 
The  architecture and the specific computation of the parameters of the network are  provided in the proof of the theorem.

\section{Simplicial Approximation Theorem Extension}\label{sec:SimpAprox}

In this section, 
we provide an extension of the Simplicial Approximation Theorem together with an explicit algorithm to compute a simplicial approximation ``as close as desired'' to a given continuous function $g:|K|\to|L|$  between the underlying spaces of
two simplicial complexes $K$ and $L$. 
The first observation is that the Simplicial Approximation Theorem (Theorem \ref{th:simaprox}) refers to any continuous function. 
Nevertheless, continuity is a property of functions between  topological spaces, not necessarily metric spaces. The next result introduces the concept of closeness between simplicial approximations and continuous functions.

\begin{proposition}\label{th:simaprox1}
 Given $\epsilon>0$ and a continuous function $g:|K|\to |L|$ between the underlying spaces of two simplicial complexes $K$ and $L$, there exist $t_1,t_2>0$ such that $\varphi_c: |\Sd^{t_1}K|\to |\Sd^{t_2}L|$ is a simplicial approximation of $g$ and $||g-\varphi_c||\le \epsilon$.
\end{proposition}

\begin{pf}
By Theorem \ref{th:barycentricepsilon}, there exists $t_2$ such that $m(\Sd^{t_2}L)\le \epsilon$. Then, by Theorem \ref{th:simaprox}, there exists $t_1$ such that $\varphi_c:|\Sd^{t_1}K|$ $\to$  $|\Sd^{t_2}L|$ is a simplicial approximation of $g$:

    $$
\begin{tikzcd}
\vert K\vert \arrow[r, "g"]       & \vert L \vert                     \\
K \arrow[u, hook] \arrow[d, "\Sd"] & L \arrow[u, hook] \arrow[d, "\Sd"] \\
\Sd^{t_1}K  \arrow[d,hook] & \Sd^{t_2}L \arrow[d,hook] \\                      \vert \Sd^{t_1}K \vert \arrow[r, "\varphi_c"]               & \vert \Sd^{t_2}L \vert      
\end{tikzcd}
$$

\noindent Besides, $||g-\varphi_c||\le \epsilon$ because $m(\Sd^{t_2}L)\le \epsilon$.
\hfill \qed
\end{pf}

\begin{algorithm2e}
  \KwIn {
  A continuous function $g:|K|\to |L|$  between the underlying spaces of two simplicial complexes $K$ and $L$, and an integer $t$ where $\Sd^t k$ satisfies the {\it star condition}:
  for each $v\in 
  \Sd^t K^{(0)}$ there exists $w\in L^{(0)}$ such that
   $g(|st(v)|)\subseteq |st(w)|$.
 }
  \KwOut{A vertex map $\varphi$ that induces simplicial approximation $\varphi_c$ of $g$.}
  \BlankLine
  \ForEach{ 
  vertex $v\in \Sd^t K^{(0)}$
  }
  {
  Choose $w\in L^{(0)}$ such that $g(|st(v)|)\subseteq |st(w)|$ and define
  $\varphi(v):=w$.
   }
 \caption{Computing a vertex map that induces a simplicial approximation}
  \label{algo-aprox}
\end{algorithm2e}

Algorithm \ref{algo-aprox} is inspired in the proof of the Simplicial Approximation Theorem given in \cite[p. 56]{edelsbrunner2010} and computes a 
vertex map $\varphi: \Sd^t K^{(0)}\to 
L^{(0)}
$ from which we can define 
the 
simplicial approximation 
 $\varphi_c:$ $|\Sd^t K|$ $\to$ $|L|$
of a continuous function $g:|K|\to |L| $.

\begin{theorem}
 Given a continuous function $g:|K|\to |L|$ and $\epsilon>0$, a two-hidden-layer feed-forward network ${\cal N}$ such that $||g-{\cal N}||\leq \epsilon$ can be explicitly defined.
\end{theorem}

\begin{pf}
By Proposition \ref{th:simaprox1}, there exists a simplicial approximation $\varphi_c$ of $g$ such that $||g-\varphi_c||\le \epsilon$, that can be computed using Algorithm~\ref{algo-aprox}. Then,
by Theorem \ref{th:simplicialmap_nn} there exists ${\cal N}_\varphi$ such that $\varphi_c={\cal N}_\varphi$. 
\hfill \qed
\end{pf}

\section{Universal Approximation Theorem Extension}\label{sec:UnivApprox}

In the previous sections, we have proved that a continuous function between triangulable 
spaces can be approximated by using the Simplicial Approximation Theorem. In this section, using the extension of the Simplicial Approximation Theorem given in Proposition \ref{th:simapproxextension}, we provide a constructive version of the Universal Approximation Theorem that approximates any continuous function (under some specific conditions) arbitrarily close. 

\begin{proposition}\label{prop:measurebarycentric}
 Let 
 $(K,\tau)$ be a finite triangulation of a metric space $X$.
 For all $\epsilon>0$ there exists $t>0$ and $\gamma>0$ such that if $m(\Sd^t K)\le \gamma$ then $\tilde{m}_{(\Sd^t K,\tau)}(X)\le \epsilon$.
\end{proposition}
\begin{pf}
Let us consider 
$a,b_0\in |K|$. 
Then, $a\in\sigma_0$ 
for some maximal simplex $\sigma_0\in K$.
If $b_0$ belongs to $\sigma_0$ then $||a-b_0||\le \delta(\sigma_0)$ and
$d_X(x,y_0)\le \tilde{\delta}(\sigma_0)$, being
$x=\tau^{-1}(a)$ and $y_0=\tau^{-1}(b_0)$.
Otherwise, we repeat the reasoning with $\Sd K$.
Now, 
$a$ and $b_0$
can belong to the same simplex  in $\Sd K$ or not. If they belong to  the same simplex in $\Sd K$, 
write 
$b_1=b_0$.
If not, take a new point 
$b_1$
such that 
$a,b_1
\in\sigma_1$ and $\sigma_1\in \Sd \sigma_0$.
Therefore, $
||a-b_1||
\le \delta(\sigma_1)\le \delta (\sigma_0)$. Besides,  $
d_X(x,y_1)
\le \tilde{\delta}(\sigma_1)\le \tilde{\delta}(\sigma_0)$
being $x=\tau^{-1}(a)$ and $y_1=\tau^{-1}(b_1)$
This process can be iterated:  $
||a-b_t||
\le \delta(\sigma_t) \le \dots \le \delta(\sigma_1)\le \delta (\sigma_0)$ and $
d_X(x,y_t)\le \tilde{\delta}(\sigma_t)\le \dots \le \tilde{\delta}(\sigma_1)\le \tilde{\delta}(\sigma_0)$,
being $x=\tau^{-1}(a)$ and $y_t=\tau^{-1}(b_t)$. By this, we have defined a sequence $\{b_i\}_{i=0}^t$ that converges to $a$.
Therefore, given $\epsilon>0$, there exists $t$ such that $d_X(x,y_t)\le \epsilon$, being
$x=\tau^{-1}(a)$ and $y_t=\tau^{-1}(b_t)$. 
Let us suppose, without loss of generality, that $\tilde{\delta}(\sigma_t)=\tilde{m}_{(\Sd^t K,\tau)}(X)$. Then, we can consider $\gamma = m(\Sd^t K)$.
\qed
\end{pf}

\begin{corollary}\label{cor:barycentricmeasure}
Given $\epsilon>0$ and a finite triangulation $(K,\tau)$ of a metric space $X$, 
there exists $t$ such that $$\tilde{m}_{(\Sd^t K,\tau)}(X)\le \epsilon.$$
\end{corollary}
\begin{pf}
By Theorem \ref{th:barycentricepsilon} there exists $t'$ such that $m(\Sd^{t'}K)$ $\le \gamma$. Then, by Proposition \ref{prop:measurebarycentric}, there exists $t$ such that $\tilde{m}_{(\Sd^t K,\tau)}(X)$ $\le \epsilon$.
\hfill \qed
\end{pf}

Given two continuous function $g_1$ and $g_2$ between two metric spaces $X$ and $Y$, we denote $\sup\{d_Y(g_1(x),g_2(x))\ | \ x\in X\}$ also by $||g_1-g_2||$.
Now, given a continuous function $g$ between two finitely triangulable metric spaces $X$ and $Y$, there exists
a  simplicial approximation $\varphi_c$ 
``arbitrarily close'' to $g$.

\begin{proposition}\label{th:simapproxextension}
Let $X$ and $Y$ be two finitely triangulable
metric
spaces, $g:X\to Y$ a continuous 
function,
and $\epsilon>0$. Then, there exist two finite triangulations $(K,\tau_K)$ and $(L,\tau_L)$ of $X$ and $Y$, respectively, 
and a simplicial approximation $\varphi_c:$ $|\Sd^{t_1}K|\to |\Sd^{t_2}L|$
such that 
$||g-\tilde{\varphi_c}||\le\epsilon$
being $\tilde{\varphi_c}=\tau^{-1}_L\circ \varphi_c\circ \tau_K$.
\end{proposition}

\begin{pf}
 First, by Corollary \ref{cor:barycentricmeasure}, there exists $t_2$ such that \newline
 $\tilde{m}_{(\Sd^{t_2}L,\tau_L)}(Y)\le \epsilon$. Next, by Theorem \ref{th:simaprox}, there exists $t_1>0$ and a vertex map $\varphi: (\Sd^{t_1}K)^{(0)}\to (\Sd^{t_2}L)^{(0)}$ such that $\varphi_c:|\Sd^{t_1}K|\to |\Sd^{t_2}L|$ is a simplicial approximation of
 $\tau_{L}\circ g\circ \tau^{-1}_{K}$. 
 Take
 into account that $|\Sd^{t_1}K|= |K|$ and $|\Sd^{t_2}L|= |L|$.
Finally, 
since 
$\tilde{m}_{(L,\tau_L)}(Y)\le \epsilon$ then
$||g-\tilde{\varphi_c}||\le\epsilon $.
Below, a diagram that schematizes the proof:
$$
\begin{tikzcd}
X \arrow[d, "\tau_{K}"] \arrow[r, "g"]   & Y \arrow[d, "\tau_{L}"]            \\
\vert K \vert \arrow[r, "\varphi_c"] & \vert L \vert                      \\
K \arrow[u, hook] \arrow[d, "\Sd"]      & L  \arrow[u, hook] \arrow[d, "\Sd"] \\
\Sd^{t_1}K   & \Sd^{t_2}L    \\
(\Sd^{t_1}K)^{(0)} \arrow[u, hook] \arrow[r, "\varphi"]      & (\Sd^{t_2}L)^{(0)}  \arrow[u, hook] 
\end{tikzcd}
$$
\hfill \qed
\end{pf}

Finally, we reach the main result of this section: Given a continuous function
$g$ between two finitely triangulable spaces $X$ and $Y$, we can obtain two finite simplicial complexes $K$ and $L$ associated to them, and a 
multi-layer feed-forward network
between the underlying spaces of $K$ and $L$  which ``approximates'' $g$.

\begin{theorem}\label{th:hidden-layer-extension}
 Given a continuous function $g:X\to Y$ between two finitely triangulable metric spaces $X$ and $Y$ and finite triangulations $(K,\tau_K)$ and $(L,\tau_L)$ of, respectively, $X$ and $Y$,
a two-hidden-layer feed-forward network $\mathcal{N}$ such that $||g-\tilde{\cal N}||\le \epsilon$, being $\tilde{\cal N}=\tau_L^{-1}\circ {\cal N} \circ \tau_K$, can be explicitly defined.
\end{theorem}
\begin{pf}
By Proposition \ref{th:simapproxextension}, there exists a simplicial approximation $\varphi_c$ 
such that 
$||g-\tilde{\varphi_c}||\leq \epsilon$.
Finally, by Theorem \ref{th:simplicialmap_nn}, there exists $\mathcal{N}$ such that $\mathcal{N} = \varphi_c$ in all the domain. \hfill \qed
\end{pf}

\section{Complexity of the Architecture of the Network}

In previous sections, we have provided a constructive approach to build neural networks to approximate continuous functions as close as desired. Now, let us study
how the ``complexity'' of the 
architecture of the neural network
increases in terms of the number of neurons in each hidden layer. 

\begin{definition}
    The complexity of a
    neural network is the maximum of the widths of its hidden layers. 
\end{definition}

First, let us observe that we can infer an upper bound for the amount of barycentric subdivisions of a simplicial complex needed to reach an specific 
mesh.

\begin{proposition}
Let us consider a finite pure simplicial complex $K$. Let $\dim(K)=n$ and let $0<\varepsilon<m(K)$.
If 
$$t \ge \frac{\log(m(K))-\log(\varepsilon)}{\log(n+1)-\log(n)} $$
then $m(\Sd^t(K))\leq  \varepsilon$.
\end{proposition}

\begin{pf}
Let us observe that $m(\Sd^t(K))\le m(K)\cdot \left(\frac{n}{n+1}\right)^t$. Then, 
$$m(\Sd^t(K))\leq\varepsilon
\Leftarrow
m(K)\cdot \left(\frac{n}{n+1}\right)^t\leq\varepsilon 
\Leftrightarrow
t\geq 
\frac{\log \left(\frac{\varepsilon}{m(K)}\right)}{\log \left(\frac{n}{n+1}\right)}.
$$
\qed
\end{pf}

Let us recall that, given two finite pure simplicial complexes 
$K$ and $L$ with, respectively, $k$ and $\ell$ maximal simplices, being, respectively, $\dim(K)=n$ and $\dim(L)=m$,
the width of  the first  and  the second hidden layer of the neural network $\mathcal{N}$ described in Theorem \ref{th:simplicialmap_nn} 
is, respectively, 
$k\cdot(n+1)$ and
$\ell\cdot(m+1)$.
Let us describe how the complexity of $\mathcal{N}$ increases with the iterated applications of the barycentric subdivisions. Let us consider, without loss of generality, that we apply one
barycentric subdivision to $K$.
Then, the width of the first hidden layer increases from $k\cdot(n+1)$ to $k \cdot(n+1)!\cdot(n+1)$.

\begin{rmk}
 Let us consider a simplicial approximation $\psi_c:|\Sd^{t_1} K|\to 
|\Sd^{t_2} L|$ of a continuous function,  being $K$ and $L$ two finite pure simplicial complexes of dimension $n$ and $m$, and with $k$ and $\ell$ maximal simplices, respectively.
The complexity of the two-hidden-layer feed-forward network ${\cal N}_{\psi} $
is
$$C(t_1,t_2)=\max\Big\{k\big((n+1)!\big)^{t_1}(n+1),\;\ell\big((m+1)!\big)^{t_2}(m+1)\Big\}.$$

\end{rmk}

We can relate the modulus of continuity of the
input function $g$ and the modulus of continuity of the simplicial approximation
$\varphi_c$ between triangulations of the input spaces. Let us recall the definition of modulus of continuity of a function.

\begin{definition}
The modulus of continuity of a continuous function $g:X\to Y$ 
between two metric spaces $X$ and $Y$ 
is given by:
$$\rho(\delta,g)=
 \sup\big\{d_Y(g(x),g(y))\ | \ d_X(x,y)\le \delta \big\}.
$$
In particular we have that:
$$d_X(x,y)\le \delta \Rightarrow d_Y(g(x),g(y))\le \rho(\delta,g).$$
\end{definition}

Now, let us study the complexity of the resulting network in terms of the mesh
of the triangulation  and on the modulus of continuity of the considered function.

\begin{theorem}
Let $g: X\to Y$ be a continuous function between two triangulable metric spaces with modulus of continuity $\rho(\delta, g)\geq 0$. Let us suppose that there exist finite triangulations $(K,\tau_K)$ and $(L,\tau_L)$ of $X$ and $Y$, respectively. Then, there exists a two-hidden-layer feed-forward network ${\cal N}$ 

such that 
$\rho(\delta,\tilde{\cal N})\leq 2\rho(\delta,g)$ and $||g-\tilde{\cal N}||\leq \frac{\rho(\delta,g)}{2}$
being $\tilde{\cal N}=\tau^{-1}_L\circ {\cal N}\circ \tau_K$.

\end{theorem}

\begin{pf}
Let $\epsilon=\frac{\rho(\delta,g)}{2}$.
By Theorem~\ref{th:hidden-layer-extension}, there exists a two-hidden-layer feed-forward network ${\cal N}$ such that 
$||g-\tilde{\cal N}||\leq \epsilon$.
Consider $x,y\in X$ such that $d_X(x,y)\leq \delta$.
then,
$$\begin{array}{l}
d_Y(\tilde{\cal N}(x),\tilde{\cal N}(y))\\
\leq  d_Y(g(x),\tilde{\cal N}(x))+
d_Y(g(x),g(y))+d_Y(g(y),\tilde{\cal N}(y))\\
\leq  2||g-\tilde{\cal N}||+\rho(\delta,g)
\leq 2\rho(\delta,g).
\end{array}
$$
\hfill \qed
\end{pf}

\section{Example}\label{sec:example}

In this section, 
we show a concrete example 
of a multi-layer feed-forward network approximating a continuous function between two triangulable spaces.The following diagram illustrates the example:
$$
\begin{tikzcd}
B^3  \arrow[r, "g"] \arrow[d, "\tau_{K}"]  & B^2   \arrow[d, "\tau_{L}"]          \\
\includegraphics[scale = 0.1]{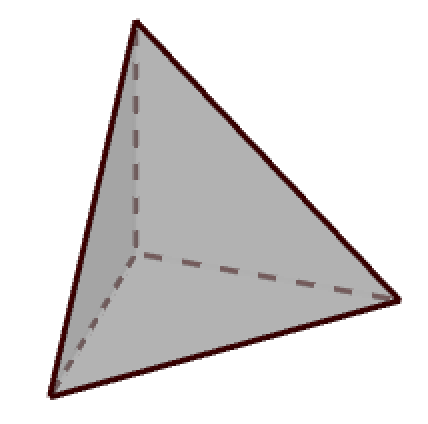}  \arrow[r, "\varphi_c"] & \includegraphics[scale = 0.1]{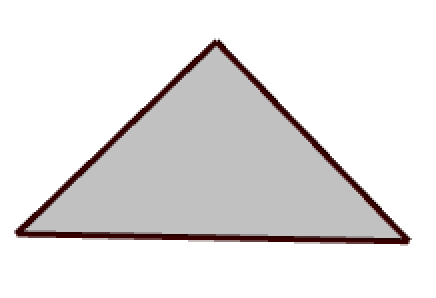}   \\
K \arrow[u, hook]  & L \arrow[u, hook] \\
K^{(0)} \arrow[u, hook] \arrow[r,"\varphi"]  & L^{(0)} \arrow[u, hook]
\end{tikzcd}
$$

\begin{figure}
    \centering
    \includegraphics[width = 0.8 \linewidth]{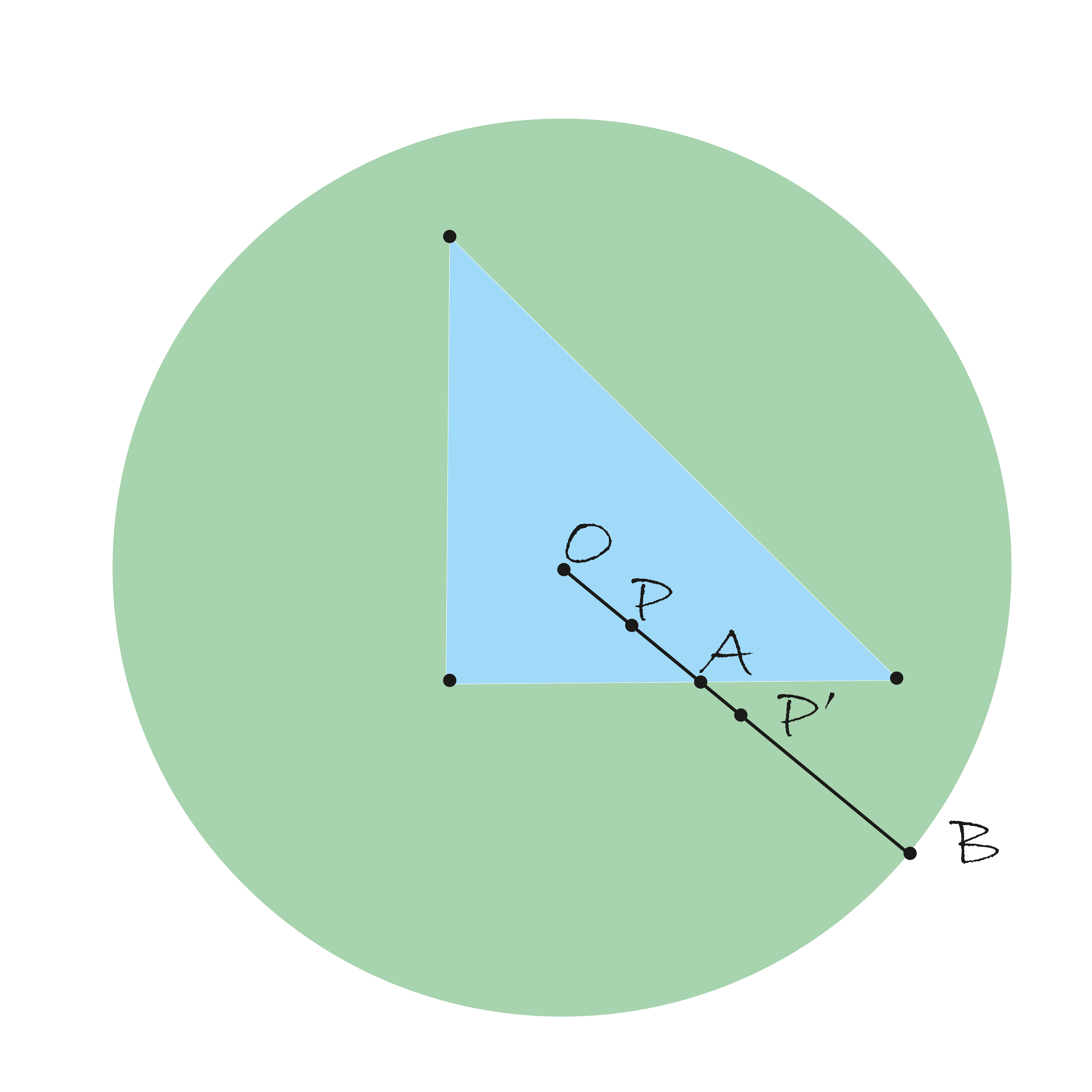}
    \caption{Geometric visualization of the computation of the parameter $\lambda$ in the homeomorphism $\tau_L$ by which $P$ is mapped to $P'$.}
    \label{fig:lambda}
\end{figure}

Let us consider the $n$-dimensional ball $B^n=\big\{x\in \mathbb{R}^n$ $| \ 1\geq ||x||\big\}$.
Then, a triangulation of $B^3$ is the simplicial complex $K$ whose maximal simplex is a a tetrahedron with set of vertices $K^{(0)}=\big\{(0,0,0)$, $(1,0,0)$, $(0,1,0)$, $(0,0,1)\big\}$ 
and a homeomorphism $\tau_K:B^3\rightarrow |K|$
whose inverse
is defined
for any point
$P\in |K|$
as follows:
$$P'=\tau^{-1}_K(P)=\lambda(P)\cdot \Big(P-\big(\frac{1}{4},\frac{1}{4},\frac{1}{4}\big)\Big)$$
where $\lambda(P)$
is 
the quotient $|\overline{OB}|/|\overline{OA}|$
being point $O=(0,0,0)$, point $A$ the intersection
of the boundary of the tetrahedron with the 
line that goes through $O$ and $P$, and point $B$ the intersection of such line with the sphere 
$S^2=\big\{x\in \mathbb{R}^3$ $| \ 1=||x||\big\}$.
See Figure~\ref{fig:lambda} for a similar homeomorphism  in the 2-dimensional case.

Besides, let us consider the continuous function
$g:B^3\rightarrow B^2$ where
$g:B^3\rightarrow B^2$ is the projection
given by $g:(P_x,P_y,P_z)\to (P_x,P_y)$
being $P=(P_x,P_y,P_z)\in B^3$. 
Besides,
$B^2$ can be triangulated by  the simplicial complex $L$ whose maximal simplex is the triangle
with set of vertices $L^{(0)}=\{(0,0)$, $(1,0)$, $(0,1)\}$ and the homeomorphism $\tau_L:B^2\to |L|$, whose inverse is 
$$P'=\tau^{-1}_L(P)=\lambda(P)\cdot \Big(P-\big(\frac{1}{4},\frac{1}{4}\big)\Big)$$
where $P\in |L|$ and $\lambda(P)$ is computed in a similar way than above.  See Figure~\ref{fig:lambda}.

Now, let us approximate $g$ with a two-hidden-layer neural network. First,
let us observe that a simplicial approximation $\varphi_c$ of
$\tau_L\circ g\circ \tau^{-1}_K$ is given by the vertex set 
$\varphi:K^{(0)}\to L^{(0)}$
defined as $\varphi((0,0,0)) = (0,0)$, $\varphi((1,0,0))=(1,0)$, $\varphi((0,1,0))=(0,1)$, and $\varphi((0,0,1))=(0,1)$. Once the simplicial approximation is computed, we can determine the specific two-hidden-layer neural network ${\cal N}_{\varphi}$
that acts equivalently to $\varphi_c$. Concretely, the architecture of ${\cal N}_{\varphi}$ is composed by an input layer with $3$ neurons, a first hidden layer with $4$ neurons, a second hidden layer with $3$ neurons, and an output layer with $2$ neurons. The weights and bias can be computed following the proof of Theorem \ref{th:simplicialmap_nn}:
$$
\begin{pmatrix}
{0} & {1} & {0} & {0} \\ {0} & {0} & {1} & {0} \\ {0} & {0} & {0} & {1} \\ {1} & {1} & {1} & {1}
\end{pmatrix}^{-1}=
\begin{pmatrix}
{-1} & {-1} & {-1} & {1} & {1} \\ {1} & {0} & {0} & {1} & {0} \\ {0} & {1} & {0} & {1} & {0} \\ {0} & {0} & {1} & {1} & {0}
\end{pmatrix}
$$
Then, 
$
W^{(1)}=\begin{pmatrix}
{-1} & {-1} & {-1}  \\ 
{1} & {0} & {0}   \\ 
{0} & {1} & {0}   \\ 
{0} & {0} & {1} 
\end{pmatrix}
$
and
$b_1 = \begin{pmatrix}
1 \\
0 \\
0\\
0
\end{pmatrix}$.

$$
W^{(2)}=
\begin{pmatrix}
{1} & {0} & {0} & {0} \\ {0} & {1} & {0} & {0} \\ {0} & {0} & {1} & {1}
\end{pmatrix}
$$
$$
W^{(3)} = 
\begin{pmatrix}
0&1&0 \\
0&0&1
\end{pmatrix}
$$

In this 
straightforward
example, the spaces
$B^3$ and  $B^2$ are approximated by just one $3$-simplex and one $2$-simplex, respectively. Therefore, $\tilde{m}_{(K,\tau_K)}(B^3)$ and $\tilde{m}_{(L,\tau_L)}(B^2)$ are upper bounded by the size of $B^3$ and $B^2$, respectively. If we want  a better approximation  to $g$, we should apply the barycentric subdivision to $K$ and $L$, respectively. Doing that,
 we would  obtain six maximal $2$-simplices in $L$, and  sixteen maximal $3$-simplices in $K$. Hence, the architecture of the neural network will consist of $64$ neurons in the first hidden layer, and $18$ neurons in the second hidden layer.

\section{Conclusion}\label{sec:conclusions}
In this paper, we have provided an effective method to build a multi-layer feed-forward network which approximates a continuous function between triangulable spaces. The main contribution of the paper is the proof that the weights can be exactly 
computed without any training process. 
Although the homeomorphisms between the triangulable spaces and the simplicial complexes can be hard to find and  the classic theorem for approximations through
neural networks is valid for compact sets and our result is only valid for triangulable spaces,
most of the real-world problems are covered by our result 
and, therefore, approximations to continuous functions 
through
neural networks can  effectively be built. Besides,
our method can be considered a suitable and powerful tool for the approximation 
of continuous functions on triangulable
spaces. Two of the 
main advantages of the proposed method are: (1)
knowing
a priori how many hidden neurons are needed 
to reach the desired accuracy; and (2) 
no need for a training process.

%\appendix
%\section{My Appendix}
%Appendix sections are coded under \verb+\appendix+.

%\verb+\printcredits+ command is used after appendix sections to list 
%author credit taxonomy contribution roles tagged using \verb+\credit+ 
%in frontmatter.

\printcredits

%% Loading bibliography style file
%\bibliographystyle{model1-num-names}
\bibliographystyle{cas-model2-names}

% Loading bibliography database
\bibliography{cas-refs}

\begin{thebibliography}{24}
\expandafter\ifx\csname natexlab\endcsname\relax\def\natexlab#1{#1}\fi
\providecommand{\url}[1]{\texttt{#1}}
\providecommand{\href}[2]{#2}
\providecommand{\path}[1]{#1}
\providecommand{\DOIprefix}{doi:}
\providecommand{\ArXivprefix}{arXiv:}
\providecommand{\URLprefix}{URL: }
\providecommand{\Pubmedprefix}{pmid:}
\providecommand{\doi}[1]{\href{http://dx.doi.org/#1}{\path{#1}}}
\providecommand{\Pubmed}[1]{\href{pmid:#1}{\path{#1}}}
\providecommand{\bibinfo}[2]{#2}
\ifx\xfnm\relax \def\xfnm[#1]{\unskip,\space#1}\fi
%Type = Book
\bibitem[{Ayala et~al.(2002)Ayala, Dom{\'\i}nguez and
  Quintero}]{ayala2002elementos}
\bibinfo{author}{Ayala, R.}, \bibinfo{author}{Dom{\'\i}nguez, E.},
  \bibinfo{author}{Quintero, A.}, \bibinfo{year}{2002}.
\newblock \bibinfo{title}{Elementos de la teor{\'\i}a de homolog{\'\i}a
  cl{\'a}sica}.
\newblock Ciencias (Universidad de Sevilla), \bibinfo{publisher}{Secretariado
  de Publicaciones, Universidad de Sevilla}.
\newblock \URLprefix \url{https://books.google.es/books?id=CAOjRFAMJFUC}.
%Type = Book
\bibitem[{Boissonnat et~al.(2018)Boissonnat, Chazal and
  Yvinec}]{boissonnat2018geometric}
\bibinfo{author}{Boissonnat, J.}, \bibinfo{author}{Chazal, F.},
  \bibinfo{author}{Yvinec, M.}, \bibinfo{year}{2018}.
\newblock \bibinfo{title}{Geometric and Topological Inference}.
\newblock Cambridge Texts in Applied Mathematics, \bibinfo{publisher}{Cambridge
  University Press}.
\newblock \URLprefix \url{https://books.google.es/books?id=0rBoDwAAQBAJ}.
%Type = Inproceedings
\bibitem[{Cohen et~al.(2016)Cohen, Sharir and
  Shashua}]{DBLP:conf/colt/CohenSS16}
\bibinfo{author}{Cohen, N.}, \bibinfo{author}{Sharir, O.},
  \bibinfo{author}{Shashua, A.}, \bibinfo{year}{2016}.
\newblock \bibinfo{title}{On the expressive power of deep learning: {A} tensor
  analysis}, in: \bibinfo{editor}{Feldman, V.}, \bibinfo{editor}{Rakhlin, A.},
  \bibinfo{editor}{Shamir, O.} (Eds.), \bibinfo{booktitle}{Proceedings of the
  29th Conference on Learning Theory, {COLT} 2016, New York, USA, June 23-26,
  2016}, \bibinfo{publisher}{JMLR.org}. pp. \bibinfo{pages}{698--728}.
\newblock \URLprefix \url{http://proceedings.mlr.press/v49/cohen16.html}.
%Type = Inproceedings
\bibitem[{Cohen and Shashua(2016)}]{DBLP:conf/icml/CohenS16}
\bibinfo{author}{Cohen, N.}, \bibinfo{author}{Shashua, A.},
  \bibinfo{year}{2016}.
\newblock \bibinfo{title}{Convolutional rectifier networks as generalized
  tensor decompositions}, in: \bibinfo{editor}{Balcan, M.},
  \bibinfo{editor}{Weinberger, K.Q.} (Eds.), \bibinfo{booktitle}{Proceedings of
  the 33nd International Conference on Machine Learning, {ICML} 2016, New York
  City, NY, USA, June 19-24, 2016}, \bibinfo{publisher}{JMLR.org}. pp.
  \bibinfo{pages}{955--963}.
\newblock \URLprefix \url{http://proceedings.mlr.press/v48/cohenb16.html}.
%Type = Article
\bibitem[{Cybenko(1989)}]{DBLP:journals/mcss/Cybenko89}
\bibinfo{author}{Cybenko, G.}, \bibinfo{year}{1989}.
\newblock \bibinfo{title}{Approximation by superpositions of a sigmoidal
  function}.
\newblock \bibinfo{journal}{Mathematics of Control Signal and Systems}
  \bibinfo{volume}{2}, \bibinfo{pages}{303--314}.
\newblock \URLprefix \url{https://doi.org/10.1007/BF02551274},
  \DOIprefix\doi{10.1007/BF02551274}.
%Type = Inproceedings
\bibitem[{Delalleau and Bengio(2011)}]{DBLP:conf/nips/DelalleauB11}
\bibinfo{author}{Delalleau, O.}, \bibinfo{author}{Bengio, Y.},
  \bibinfo{year}{2011}.
\newblock \bibinfo{title}{Shallow vs. deep sum-product networks}, in:
  \bibinfo{editor}{Shawe{-}Taylor, J.}, \bibinfo{editor}{Zemel, R.S.},
  \bibinfo{editor}{Bartlett, P.L.}, \bibinfo{editor}{Pereira, F.C.N.},
  \bibinfo{editor}{Weinberger, K.Q.} (Eds.), \bibinfo{booktitle}{Advances in
  Neural Information Processing Systems 24: 25th Annual Conference on Neural
  Information Processing Systems 2011. Proceedings of a meeting held 12-14
  December 2011, Granada, Spain}, pp. \bibinfo{pages}{666--674}.
\newblock \URLprefix
  \url{http://papers.nips.cc/paper/4350-shallow-vs-deep-sum-product-networks}.
%Type = Book
\bibitem[{Edelsbrunner and Harer(2010)}]{edelsbrunner2010}
\bibinfo{author}{Edelsbrunner, H.}, \bibinfo{author}{Harer, J.},
  \bibinfo{year}{2010}.
\newblock \bibinfo{title}{Computational Topology - an Introduction.}
\newblock \bibinfo{publisher}{American Mathematical Society}.
%Type = Book
\bibitem[{Geoghegan(2010)}]{geoghegan2010topological}
\bibinfo{author}{Geoghegan, R.}, \bibinfo{year}{2010}.
\newblock \bibinfo{title}{Topological Methods in Group Theory}.
\newblock Graduate Texts in Mathematics, \bibinfo{publisher}{Springer New
  York}.
\newblock \URLprefix \url{https://books.google.ch/books?id=MEKZcQAACAAJ}.
%Type = Article
\bibitem[{Guliyev and Ismailov(2018)}]{GULIYEV2018296}
\bibinfo{author}{Guliyev, N.J.}, \bibinfo{author}{Ismailov, V.E.},
  \bibinfo{year}{2018}.
\newblock \bibinfo{title}{On the approximation by single hidden layer
  feedforward neural networks with fixed weights}.
\newblock \bibinfo{journal}{Neural Networks} \bibinfo{volume}{98},
  \bibinfo{pages}{296 -- 304}.
\newblock \URLprefix
  \url{http://www.sciencedirect.com/science/article/pii/S0893608017302927},
  \DOIprefix\doi{https://doi.org/10.1016/j.neunet.2017.12.007}.
%Type = Misc
\bibitem[{Hanin and Sellke(2017)}]{1710.11278}
\bibinfo{author}{Hanin, B.}, \bibinfo{author}{Sellke, M.},
  \bibinfo{year}{2017}.
\newblock \bibinfo{title}{Approximating continuous functions by relu nets of
  minimal width}.
\newblock \href{http://arxiv.org/abs/arXiv:1710.11278}{\tt
  arXiv:arXiv:1710.11278}.
%Type = Book
\bibitem[{Hatcher(2002)}]{Hatcher}
\bibinfo{author}{Hatcher, A.}, \bibinfo{year}{2002}.
\newblock \bibinfo{title}{Algebraic topology}.
\newblock \bibinfo{publisher}{Cambridge University Press},
  \bibinfo{address}{Cambridge}.
%Type = Book
\bibitem[{Haykin(1999)}]{haykin99a}
\bibinfo{author}{Haykin, S.}, \bibinfo{year}{1999}.
\newblock \bibinfo{title}{Neural Networks: A Comprehensive Foundation}.
\newblock \bibinfo{publisher}{Prentice Hall}.
%Type = Article
\bibitem[{Hornik(1991)}]{Hornik:1991:ACM:109691.109700}
\bibinfo{author}{Hornik, K.}, \bibinfo{year}{1991}.
\newblock \bibinfo{title}{Approximation capabilities of multilayer feedforward
  networks}.
\newblock \bibinfo{journal}{Neural Networks} \bibinfo{volume}{4},
  \bibinfo{pages}{251--257}.
\newblock \URLprefix \url{http://dx.doi.org/10.1016/0893-6080(91)90009-T},
  \DOIprefix\doi{10.1016/0893-6080(91)90009-T}.
%Type = Article
\bibitem[{Hornik et~al.(1989)Hornik, Stinchcombe and White}]{hornik89a}
\bibinfo{author}{Hornik, K.}, \bibinfo{author}{Stinchcombe, M.},
  \bibinfo{author}{White, H.}, \bibinfo{year}{1989}.
\newblock \bibinfo{title}{Multilayer feedforward networks are universal
  approximators}.
\newblock \bibinfo{journal}{Neural Networks} \bibinfo{volume}{2},
  \bibinfo{pages}{356--366}.
%Type = Inproceedings
\bibitem[{Kileel et~al.(2019)Kileel, Trager and
  Bruna}]{DBLP:conf/nips/KileelTB19}
\bibinfo{author}{Kileel, J.}, \bibinfo{author}{Trager, M.},
  \bibinfo{author}{Bruna, J.}, \bibinfo{year}{2019}.
\newblock \bibinfo{title}{On the expressive power of deep polynomial neural
  networks}, in: \bibinfo{editor}{Wallach, H.M.}, \bibinfo{editor}{Larochelle,
  H.}, \bibinfo{editor}{Beygelzimer, A.}, \bibinfo{editor}{d'Alch{\'{e}}{-}Buc,
  F.}, \bibinfo{editor}{Fox, E.B.}, \bibinfo{editor}{Garnett, R.} (Eds.),
  \bibinfo{booktitle}{Advances in Neural Information Processing Systems 32:
  Annual Conference on Neural Information Processing Systems 2019, NeurIPS
  2019, 8-14 December 2019, Vancouver, BC, Canada}, pp.
  \bibinfo{pages}{10310--10319}.
\newblock \URLprefix
  \url{http://papers.nips.cc/paper/9219-on-the-expressive-power-of-deep-polynomial-neural-networks}.
%Type = Article
\bibitem[{Liang and Srikant(2016)}]{DBLP:journals/corr/LiangS16}
\bibinfo{author}{Liang, S.}, \bibinfo{author}{Srikant, R.},
  \bibinfo{year}{2016}.
\newblock \bibinfo{title}{Why deep neural networks?}
\newblock \bibinfo{journal}{CoRR} \bibinfo{volume}{abs/1610.04161}.
\newblock \URLprefix \url{http://arxiv.org/abs/1610.04161},
  \href{http://arxiv.org/abs/1610.04161}{\tt arXiv:1610.04161}.
%Type = Inproceedings
\bibitem[{Lu et~al.(2017)Lu, Pu, Wang, Hu and Wang}]{DBLP:conf/nips/LuPWH017}
\bibinfo{author}{Lu, Z.}, \bibinfo{author}{Pu, H.}, \bibinfo{author}{Wang, F.},
  \bibinfo{author}{Hu, Z.}, \bibinfo{author}{Wang, L.}, \bibinfo{year}{2017}.
\newblock \bibinfo{title}{The expressive power of neural networks: {A} view
  from the width}, in: \bibinfo{editor}{Guyon, I.}, \bibinfo{editor}{von
  Luxburg, U.}, \bibinfo{editor}{Bengio, S.}, \bibinfo{editor}{Wallach, H.M.},
  \bibinfo{editor}{Fergus, R.}, \bibinfo{editor}{Vishwanathan, S.V.N.},
  \bibinfo{editor}{Garnett, R.} (Eds.), \bibinfo{booktitle}{Advances in Neural
  Information Processing Systems 30: Annual Conference on Neural Information
  Processing Systems 2017, 4-9 December 2017, Long Beach, CA, {USA}}, pp.
  \bibinfo{pages}{6232--6240}.
\newblock \URLprefix
  \url{http://papers.nips.cc/paper/7203-the-expressive-power-of-neural-networks-a-view-from-the-width}.
%Type = Article
\bibitem[{Martens and Medabalimi(2014)}]{MartensM14}
\bibinfo{author}{Martens, J.}, \bibinfo{author}{Medabalimi, V.},
  \bibinfo{year}{2014}.
\newblock \bibinfo{title}{On the expressive efficiency of sum product
  networks.}
\newblock \bibinfo{journal}{CoRR} \bibinfo{volume}{abs/1411.7717}.
\newblock \URLprefix
  \url{http://dblp.uni-trier.de/db/journals/corr/corr1411.html#MartensM14}.
%Type = Book
\bibitem[{Munkres(1984)}]{munkres}
\bibinfo{author}{Munkres, J.R.}, \bibinfo{year}{1984}.
\newblock \bibinfo{title}{Elements of algebraic topology.}
\newblock \bibinfo{publisher}{Addison-Wesley}.
%Type = Article
\bibitem[{Nguyen et~al.(2018)Nguyen, Mukkamala and
  Hein}]{DBLP:journals/corr/abs-1803-00094}
\bibinfo{author}{Nguyen, Q.}, \bibinfo{author}{Mukkamala, M.C.},
  \bibinfo{author}{Hein, M.}, \bibinfo{year}{2018}.
\newblock \bibinfo{title}{Neural networks should be wide enough to learn
  disconnected decision regions}.
\newblock \bibinfo{journal}{CoRR} \bibinfo{volume}{abs/1803.00094}.
\newblock \URLprefix \url{http://arxiv.org/abs/1803.00094},
  \href{http://arxiv.org/abs/1803.00094}{\tt arXiv:1803.00094}.
%Type = Misc
\bibitem[{Poon and Domingos(2012)}]{poon2012sumproduct}
\bibinfo{author}{Poon, H.}, \bibinfo{author}{Domingos, P.},
  \bibinfo{year}{2012}.
\newblock \bibinfo{title}{Sum-product networks: A new deep architecture}.
\newblock \href{http://arxiv.org/abs/1202.3732}{\tt arXiv:1202.3732}.
%Type = Inproceedings
\bibitem[{Safran and Shamir(2017)}]{pmlr-v70-safran17a}
\bibinfo{author}{Safran, I.}, \bibinfo{author}{Shamir, O.},
  \bibinfo{year}{2017}.
\newblock \bibinfo{title}{Depth-width tradeoffs in approximating natural
  functions with neural networks}, in: \bibinfo{editor}{Precup, D.},
  \bibinfo{editor}{Teh, Y.W.} (Eds.), \bibinfo{booktitle}{Proceedings of the
  34th International Conference on Machine Learning},
  \bibinfo{publisher}{PMLR}, \bibinfo{address}{International Convention Centre,
  Sydney, Australia}. pp. \bibinfo{pages}{2979--2987}.
\newblock \URLprefix \url{http://proceedings.mlr.press/v70/safran17a.html}.
%Type = Inproceedings
\bibitem[{Sun et~al.(2016)Sun, Chen, Wang, Liu and
  Liu}]{DBLP:conf/aaai/SunCWLL16}
\bibinfo{author}{Sun, S.}, \bibinfo{author}{Chen, W.}, \bibinfo{author}{Wang,
  L.}, \bibinfo{author}{Liu, X.}, \bibinfo{author}{Liu, T.},
  \bibinfo{year}{2016}.
\newblock \bibinfo{title}{On the depth of deep neural networks: {A} theoretical
  view}, in: \bibinfo{editor}{Schuurmans, D.}, \bibinfo{editor}{Wellman, M.P.}
  (Eds.), \bibinfo{booktitle}{Proceedings of the Thirtieth {AAAI} Conference on
  Artificial Intelligence, February 12-17, 2016, Phoenix, Arizona, {USA.}},
  \bibinfo{publisher}{{AAAI} Press}. pp. \bibinfo{pages}{2066--2072}.
\newblock \URLprefix
  \url{http://www.aaai.org/ocs/index.php/AAAI/AAAI16/paper/view/12073}.
%Type = Article
\bibitem[{Telgarsky(2016)}]{DBLP:journals/corr/Telgarsky16}
\bibinfo{author}{Telgarsky, M.}, \bibinfo{year}{2016}.
\newblock \bibinfo{title}{Benefits of depth in neural networks}.
\newblock \bibinfo{journal}{CoRR} \bibinfo{volume}{abs/1602.04485}.
\newblock \URLprefix \url{http://arxiv.org/abs/1602.04485},
  \href{http://arxiv.org/abs/1602.04485}{\tt arXiv:1602.04485}.

\end{thebibliography}

\vfill\eject

\bio{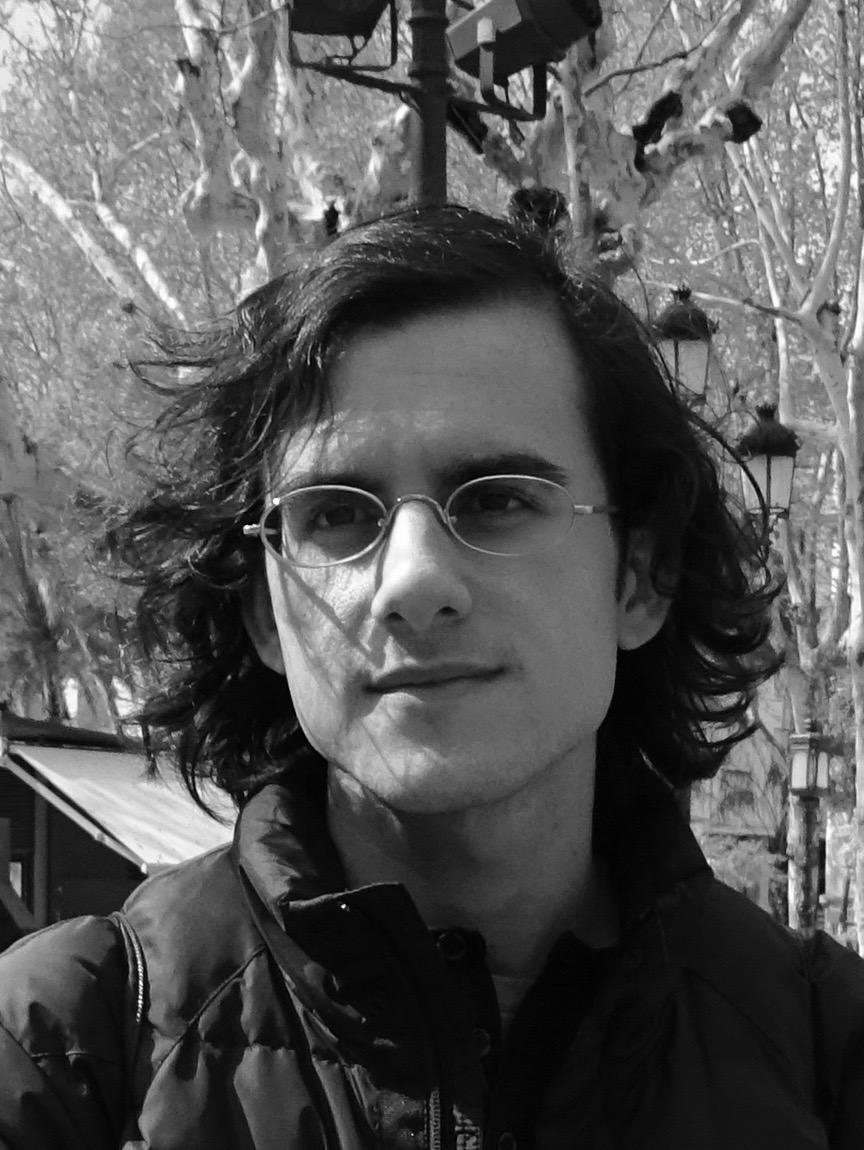}
{\bf Eduardo Paluzo-Hidalgo}
He received the degree in Mathematics in 2017 and the MSc in Logic, Computation, and Artificial Intelligence in 2018 from the University of Seville. He is currently a PhD student and researcher in the Department of Applied Math I at the University of Seville, Spain. He is a member of the Combinatorial Image Analysis group. His research interests are primarily machine learning and computational topology.
\endbio

\bio{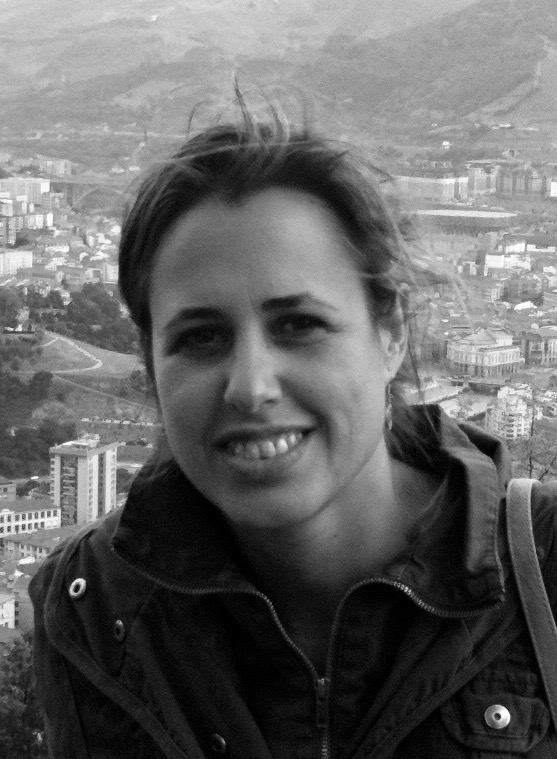}
{\bf Rocio Gonzalez-Diaz}
{
Associate Professor in the Department of Applied Math I at the University of Seville since 2004. Head of the Computational Image Analysis group since 2010. Principal Researcher of the Spanish project Computational Algebraic Topology for Computer Vision (MTM2015-67072-P Project) from 01-2016 to 12-2018.} 
\endbio

\vspace{5mm}

\bio{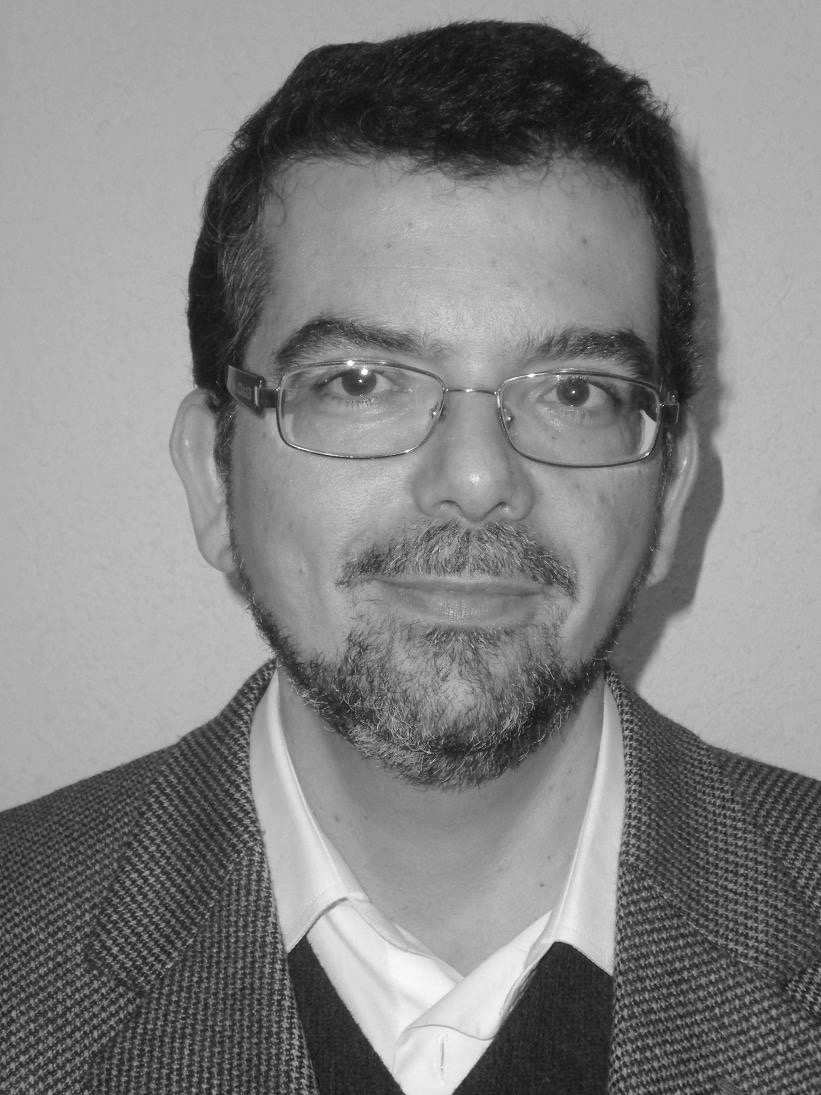}
{\bf Miguel A. Guti\'errez-Naranjo}
{Associate professor in the Department of Computer Science and Artificial Intelligence at the University of Seville, Spain. His research interest includes topics related to Artificial Intelligence and Natural Computing, both from a theoretical and practical point of view. He has coauthored more than 40 scientific papers in these areas.}
\endbio

\end{document}